\documentclass{article}
\usepackage{dcase2016,amsmath,graphicx,url,times,hyperref}

\title{FULLY DNN-BASED MULTI-LABEL REGRESSION FOR AUDIO TAGGING}

\name{Yong Xu\sthanks{This work was supported by the Engineering and Physical Sciences Research Council (EPSRC) of the UK under the grant EP/N014111/1.}, Qiang Huang\sthanks{The first author and the second author have equal contribution for this paper.}, Wenwu Wang, Philip J. B. Jackson, Mark D. Plumbley}
\address{Centre for Vision, Speech and Signal Processing, University of Surrey, UK\\
	\text{\{yong.xu, q.huang, w.wang, p.jackson, m.plumbley\}@surrey.ac.uk{\scriptsize }}
}

\begin{document}

\ninept
\maketitle

\begin{sloppy}

\begin{abstract}
Acoustic event detection for content analysis in most cases relies on a lot of labeled data. However, manually annotating data is a time-consuming task, which thus makes few annotated resources available so far. Unlike audio event detection, automatic audio tagging, a multi-label acoustic event classification task, only relies on weakly labeled data. This is highly desirable to some practical applications using audio analysis. In this paper we propose to use a fully deep neural network (DNN) framework to handle the multi-label classification task in a regression way. Considering that only chunk-level rather than frame-level labels are available, the whole or almost whole frames of the chunk were fed into the DNN to perform a multi-label regression for the expected tags. The fully DNN, which is regarded as an encoding function, can well map the audio features sequence to a multi-tag vector. A deep pyramid structure was also designed to extract more robust high-level features related to the target tags. Further improved methods were adopted, such as the Dropout and background noise aware training, to enhance its generalization capability for new audio recordings in mismatched environments. Compared with the conventional Gaussian Mixture Model (GMM) and support vector machine (SVM) methods, the proposed fully DNN-based method could well utilize the long-term temporal information with the whole chunk as the input. The results show that our approach obtained a 15\% relative improvement compared with the official GMM-based method of DCASE 2016 challenge.

\end{abstract}


\section{Introduction}
\label{sec:intro}

Due to the use of smart mobile devices in recent years, huge amounts of
multimedia data are generated and uploaded to the internet everyday. 
These data, such as music, field sounds, broadcast news, and
television shows, contain sounds from a wide variety of sources.
The need for analyzing these sounds has been now increased 
as it is useful, e.g., for automatic tagging in audio indexing, automatic sound analysis for
audio segmentation or audio context classification.
Although supervised approaches have proved to be effective in
many applications, their effectiveness relies heavily on the
quantity and quality of the training data. 
Moreover, manually labeling a large amount 
of data is very time-consuming. 
To handle this problem, two types of methods have been developed.
One is to convert low-level acoustic features into
``bag of audio words'' using unsupervised learning methods \cite{chen2010,riley2008, shao2004, cai2005,Sainath2007}.
The second type of methods is based on only weakly labeled data \cite{DBLP:journals/corr/0003R16}, e.g. 
audio tagging. It is clear that tagging audio chunks
needs much less time compared to precisely locating event
boundaries within recordings. This will certainly improve
tractability of obtaining manual annotations for large databases. In this paper, we will focus on the audio tagging task.


To overcome the lack of annotated training data,
Multiple Instance Learning (MIL) is proposed in \cite{dietterich1997} as a variation 
of supervised learning for problems with incomplete knowledge about labels of training examples.
It aims to classify sets of instances instead of recognizing single instances.
Following this work, Andrews \emph{et al.} \cite{andrew2003} proposed a new formulation of 
MIL as a maximum margin problem, which had led to some further work 
\cite{mandel2008, briggs2012, carbonetto2008,cheny2006, ulges2008}
in audio and video processing using weakly labeled data. 
Mandel and Ellis in \cite{mandel2008} used clip-level tags to derive tags at the track, album, 
and artist granularities by formulating a number of music information related
multiple-instance learning tasks and evaluated two MIL based algorithms
on them.
In \cite{phan2015}, Phan \textit{et al.} used event-driven MIL to learn the key evidences for event detection. Recently, \cite{DBLP:journals/corr/0003R16} also presented a SVM based MIL system for audio tagging and event detection.
GMM, as a common model, was used as the official baseline method in DCASE 2016 for audio tagging. More details can be found in \cite{dcase_t4}.

Although the methods mentioned above have led to
some useful results in detection and analysis of audio data,
most of them ignored possible
relationships of any contextual information and only focused on
training the model for each single event class independently.
To better use the data with weak labels, 
our work will utilize the whole or almost whole frames of the observed chunk as the input of a fully deep neural network to make a mapping from an audio feature sequence to a multi-tag vector. 

Recently, deep learning technologies have obtained great successes in speech, image and video fields \cite{xu2014experimental, xu2015regression, hinton2012deep, krizhevsky2012imagenet} since Hinton and Salakhutdinov showed the insights using a greedy layer-wise unsupervised learning procedure to train a deep model in 2006 \cite{hinton2006reducing}. The deep learning methods were also investigated for related tasks, like acoustic scene classification \cite{petetin2015deep} and acoustic event detection \cite{cakir2015polyphonic}. And better performance could be obtained in these tasks. For music tagging task, \cite{dieleman2014end, choi2016automatic} have also demonstrated the superiority of deep learning methods. However, to the best of our knowledge, the deep learning based methods
have not been used for environmental audio tagging which is a newly proposed task in \text{DCASE} 2016 challenge based on the CHiME-home dataset \cite{foster2015}. For the audio tagging task, only the chunk-level instead of frame-level labels were available. Furthermore, multiple instances could happen simultaneously, for example, the \textit{child speech} could exist with \textit{TV sound} for several seconds. Hence, a good way is to feed the DNN with the whole frames of the chunk to predict the multiple tags in the output.

In this paper, we propose a fully DNN-based method,
which can well utilize the long-term temporary information, to map the whole sequence of audio features into a multi-tag vector. The fully neural network structure was also successfully used in image segmentation \cite{long2015fully}. To get a better prediction of the tags, a deep pyramid structure is designed with gradually shrinked size of layers. This deep pyramid structure can reduce the non-correlated interferences in the whole audio features while focusing on extracting the robust high-level features related to the target tags. Dropout \cite{dahl2013improving} and background noise aware training \cite{xu2014dynamic} are adopted to further improve the tagging performance in the DNN-based framework.

The rest of the paper is organized as follows. In section \ref{sec:related_work}, we will introduce 
the related work using GMM and SVM based MIL in detail, and depict our DNN based
framework in section \ref{sec:fully_dnn}. The data description and experimental setup
will be given in section \ref{sec:exp}. We will show
the related results and discussions in section \ref{sec:results}, and finally draw a conclusion in section \ref{sec:conclusions}.

\section{Related Work}
\label{sec:related_work}
Two baseline methods compared in our work are briefly summarized below.
\subsection{Audio Tagging using Gaussian Mixture Models}\label{subsec:GMM}
Gaussian Mixture Models (GMMs) are a commonly used generative
classifier. A GMM is parametrized in 
$\Theta=\{ \omega_m, \mu_m, \Sigma_m \}, m = \{1,\cdots ,M\}$,
where $M$ is the number of mixtures and $w_m$ is the weight of the $m$-th
mixture component. 

To implement multi-label classification with simple event tags, 
a binary classifier
is built associating with each audio event class in the training step. 
For a specific event class, all audio frames in an audio chunk labeled with this event
are categorized into a positive class, whereas the remaining features are categorized into 
a negative class.  On the classification stage, given an audio
chunk $C_i$, the likelihoods of each audio frame
$x_{ij}, (j \in \{1 \cdots L_{C_i}\})$  are calculated for the two class models, respectively.
%
Given audio event class $k$ and chunk $C_i$, the classification score $S_{C_{ik}}$
is obtained as log-likelihood ratio:

\begin{equation}
 S_{C_{ik}} = \sum_j log(f(x_{ij}, \Theta_{pos})) - \sum_j log(f(x_{ij}, \Theta_{neg}))
\end{equation}

\subsection{Audio Tagging using Multiple Instance SVM }\label{subsec:MIL}

Multiple instance learning is described in terms of bags $\textbf{B}$. 
The $j$th instance in the $i$th bag, $B_i$, is defined as $x_{ij}$ 
where $j \in I=\{1 \cdots l_i\}$, and $l_i$ is the number of instances in $B_i$.
$B_i$'s label is $Y_i \in \{ -1, 1 \}$. 
If $Y_i = -1$, then $x_{ij} = -1$ for all $j$. 
If $ Y_i = 1$, then at least
one instance $x_{ij} \in B_i$ is a positive example of the underlying concept \cite{andrew2003}.

As MI-SVM is the bag-level MIL support vector machine to maximize the bag margin,
we define the functional margin of a bag with respect to a hyper-plane as:
\begin{equation}
  \gamma_i = Y_i \max_{j \in I} (\langle \textbf{w},\textbf{x}_{ij} \rangle +b) 
\end{equation}
Using the above notion, MI-SVM can be defined as:

\begin{equation}
  \min_{\textbf{w},b,\xi} \dfrac{1}{2}\Vert \textbf{w}^2 \Vert + A\sum_i\xi_i
\end{equation} 
~~~~~~~~subject~~ to:$~~~~~~\forall_i: \gamma_i \geq 1-\xi_i $,~~~$\xi_i \geq 0 $

where $\textbf{w}$ is weight vector, $b$ is bias, $\xi$ is
margin violation, and $A$ is a regularization parameter.
   
Classification with MI-SVM proceeds in two steps. 
In the first step, $\textbf{x}_i$ is initialized as the centroid
for every positive bag $B_i$ as follows

\begin{equation}
 \overline{\textbf{x}}_i = \sum_{j \in I} \textbf{x}_{ij}/l_i 
\end{equation}
The second step is an iterative procedure in order to
optimize the parameters. 

Firstly, $\textbf{w}$ and $b$ are computed for the data set 
with positive samples $\{x_I: Y_i=1\}$.

Secondly, we compute

$~~~~~~~~~~~~~~~~~~~~~~~~f_{ij} = \langle\textbf{w}, \textbf{x}_{ij}\rangle +b$, $~~~~~\textbf{x}_{ij} \in \textbf{B}_i$ 

Thirdly, we change $\overline{\textbf{x}}_i$ by

$~~~~~~~~~~~~~~~~~~~~~~~~\overline{\textbf{x}}_i = \textbf{x}_j$

$~~~~~~~~~~~~~~~~~~~~~~~~j=\arg\max_{j \in I} f_{ij}, \forall I, Y_I=1$ 

The iteration in this step will stop when there is no change of $\overline{\textbf{x}}_i$,
The optimized parameters will be used for test.

\section{Proposed Fully DNN-based audio tagging}
\label{sec:fully_dnn}
DNN is a non-linear multi-layer model for extracting robust features related to a specific classification \cite{hinton2012deep} or regression \cite{xu2015regression} task. The objective of the audio tagging task is to perform multi-label classification on audio chunks (i.e. assign zero or more labels to each audio chunk of a length e.g. four seconds in our experiments). This chunk only has utterance-level labels without frame-level labels. Multiple events happen at many particular frames. Hence, the common frame-level cross entropy based loss function can not be adopted. We propose a method to encode the whole or almost whole chunk.
\label{sec:DNN}
\begin{figure}[t]
	\centering
	\centerline{\includegraphics[width=\columnwidth]{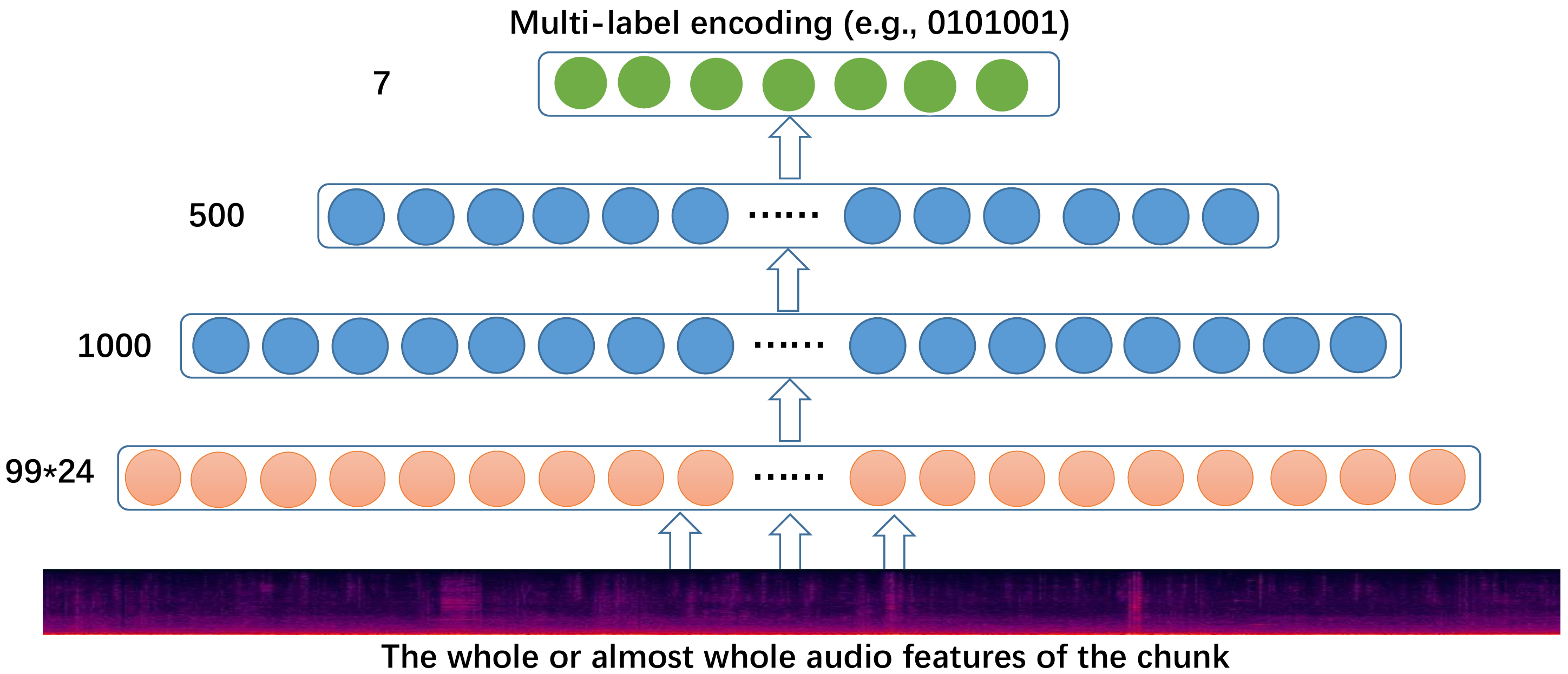}}
	\caption{Fully DNN-based audio tagging framework using the deep pyramid structure.}
	\label{fig:dnn_tag}
\end{figure}
\subsection{Fully DNN-based multi-label regression using sequence to sequence mapping}
\label{ssec:fully_dnn}
Fig. \ref{fig:dnn_tag} shows the proposed fully DNN-based audio tagging framework using the deep pyramid structure. 
With the proposed framework, the whole or almost whole audio features of the chunk are encoded into a vector with values $\{0,1\}$ in a regression way. Sigmoid was used as the activation function of the output layer to learn the presence probability of certain events. Minimum mean squared error (MMSE) was adopted as the objective function. A stochastic gradient descent algorithm is performed in mini-batches with multiple epochs to improve learning convergence as follows,
\begin{equation}
Er=\frac{1}{N}\sum_{n=1}^{N}\|\hat{\textbf{X}}_{n} (\textbf{Y}{_{n-\tau}^{n+\tau}},\textbf{W}, \textbf{b})-\textbf{X}_{n}\|_{2}^2
\label{DNNerrors}
\end{equation}
where $Er$ is the mean squared error, $\hat{\textbf{X}}_{n}(\textbf{Y}{_{n-\tau}^{n+\tau}},\textbf{W},\textbf{b})$ and $\textbf{X}_{n}$ denote the estimated and reference tag vector at sample index $n$, respectively, with $N$ representing the mini-batch size, $ \textbf{Y}{_{n-\tau}^{n+\tau}} $ being the input audio feature vector where the window size of context is $2*\tau+1$. It should be noted that the input window size should cover the whole or almost whole of the chunk considering that the reference tags are in chunk-level rather than frame-level labels. However, slightly relaxing the window size without covering all of the chunk frames could increase the total training samples for DNN. It can improve the performance in our experiments. $(\textbf{W}, \textbf{b})$ denoting the weight and bias parameters to be learned.
The updated estimate of $ \textbf{W}^\ell $ and $ \textbf{b}^\ell $ in the $\ell$-th layer, with a learning rate $\lambda$, can be computed iteratively as follows:
\begin{equation}
(\textbf{W}^\ell, \textbf{b}^\ell)~\leftarrow~(\textbf{W}^\ell, \textbf{b}^\ell)-\lambda\frac{\partial{Er}}{\partial{(\textbf{W}^\ell, \textbf{b}^\ell)}},~1\leq{\ell}\leq{L+1}
\end{equation}
where $L$ denotes the total number of hidden layers and $L+1$ represents the output layer.

During the learning process where the DNN can be regarded as an encoding function, the audio tags are automatically predicted. Hence the multi-label regression rather than classification can be conducted. Two additional methods are given below to improve the DNN-based audio tagging performance.

\subsection{Dropout for the over-fitting problem}
\label{ssec:dropout}
Deep learning architectures have a natural tendency towards over-fitting especially when there is little training data. This audio tagging task only has about four hours training data with imbalanced training data distribution for each type of tag. Dropout is a simple but effective way to alleviate this problem \cite{dahl2013improving}. In each training iteration, the feature value of every input unit and the activation of every hidden unit are randomly removed with a predefined probability (e.g., $ \rho $). These random perturbations effectively prevent the DNN from learning spurious dependencies. At the decoding stage, the DNN discounts all of the weights involved in the dropout training by $(1-\rho)$, regarded as a model averaging process \cite{hinton2012improving}.

A mismatch problem may also exist in this task, and testing audio segments could be totally different from existed training audio segments due to the presence of lots of background noise. Thus Dropout should be adopted to improve its robustness to generalize to variation in testing segments.

\subsection{Background noise aware training}
Different types of background noise in different recording environments could lead to the mismatch problem between the testing chunks and the training chunks. To alleviate this, we propose a simple background noise aware training (or background noise adaptation method). To enable this noise awareness, the DNN is fed with the primary audio features augmented with an estimate of the background noise. In this way, the DNN can use additional on-line background noise information to better predict the expected tags. The background noise is estimated as follows:
\begin{equation}
\textbf{V}_{n}=[\textbf{Y}_{n-\tau},...,\textbf{Y}_{n-1},\textbf{Y}_{n},\textbf{Y}_{n+1},...,\textbf{Y}_{n+\tau}, \hat{\textbf{Z}}_{n}]
\end{equation}
\begin{equation}
\hat{\textbf{Z}}_{n}=\dfrac{1}{T}\sum_{t=1}^{T}\textbf{Y}_{t}
\label{eq:nat}
\end{equation}
where the background noise $\hat{\textbf{Z}}_{n}$ is fixed over the utterance and estimated using the first $T$ frames. Although this noise estimator is simple, a similar idea was shown to be effective in DNN-based speech enhancement \cite{xu2015regression, xu2014dynamic}.

\section{experimental setup and results}
\label{sec:exp}
\subsection{DCASE2016 data set for audio tagging}
\label{ssec:data_set}
The data that we used for evaluation is the dataset of Task4 of \text{DCASE} 2016 \cite{dcase_t4}.
The audio recordings are made in a domestic environment.
The audio data are provided as 4-second chunks at two sampling rates 
(48kHz and 16kHz) with the 48kHz data in stereo and with the 16kHz data in mono.
The 16kHz recordings were obtained by downsampling the right-hand channel of the 48kHz recordings.
Each audio file corresponds to a single chunk \cite{dcase_t4}.

For each chunk, multi-label annotations were first obtained from each of the 3 annotators. 
The annotations are based on a set of 7 label classes.
A detailed description of the annotation procedure is provided in \cite{foster2015}.
%
To reduce uncertainty of the test data, the evaluation is based on those 
chunks where 2 or more annotators agreed about label presence across label classes.
Moreover, with the aim of approximating typical recording capabilities of commodity hardware,
only the monophonic audio data sampled at 16kHz are used for test.


\subsection{Experimental Setup}\label{ssec:exp_setup}

In our experiments, following the original configuration of
Task4 of DCASE 2016 \cite{dcase_t4},
we use the same five folds as the evaluation set from the given development dataset, and 
use the remain of the audio recordings for training.

We pre-process each audio chunk by 
segmenting them using a ($80$ms) sliding window with
a $40$ms hop size, and converting
each segment into 24-D MFCCs. For each 4-second chunk, 99 frames of MFCCs are obtained. A 91-frame expansion as the input instead of the total frames were found better because this relaxed input scheme can increase the total training samples. Hence the input size of DNN was 2208 with 91-frame MFCCs and also the appended noise vector. One hidden layer with 1000 units and the second hidden layer with 500 units were used to construct a pyramid structure. Seven sigmoid outputs were adopted to predict the seven tags. The learning rate was 0.005. Momentum was set to be 0.9. The dropout rates for input layer and hidden layer were 0.1 and 0.2, respectively. The mini-batch size was 3. $T$ in Equation \ref{eq:nat} was 6. It should be noted that the remaining 2432 chunks without `strongly agreement' labels in the development dataset were also added into the DNN training considering that DNN has a better fault-tolerant capability. Meanwhile, these 2432 chunks without `strongly agreement' labels were also added into the training data for GMM and SVM training.

For a comparison, we also ran two baselines
using GMMs and the MI-SVM mentioned
in Section \ref{sec:related_work}. For the GMM
based method, the number of mixture components is 8. 
Since the GMM based baseline focuses on computing frame-level
likelihoods and MI-SVM prefers to instance-level scores,
the sliding window and hop size set for the two baselines
are different.
The GMM based baseline uses a 20ms sliding window with 10ms hop size,
while the sliding window and hop size for MI-SVM are set to be 400ms and 200ms, respectively.  
To handle audio tagging with MI-SVM, each audio recording 
will be viewed as a bag and its shorter segments obtained by a sliding
window can be treated as an instance. 
To accelerate computation, we use linear function kernel in our experiments.

To evaluate the effectiveness of our approach, as compared with
the two baselines, we use equal error rate (EER) as a metric.
EER is defined as the point of the graph of false negative rate (FNR) versus false
positive rate (FPR) \cite{murphy2012} \\

$~~~~~~~~~~~~~~~~FNR = \dfrac{\#false ~negative}{ \#positive}$\\
\vspace{2mm}\\
$~~~~~~~~~~~~~~~~~~~~~~~~FPR = \dfrac{\#false ~positive}{\#negative}$\\
EERs are computed individually for each evaluation,
and we then average the obtained EERs across the five evaluations
to get the final performance.

\section{Results and discussions}
\label{sec:results}

\begin{figure}[t]
	\centering
	\includegraphics[scale=0.35]{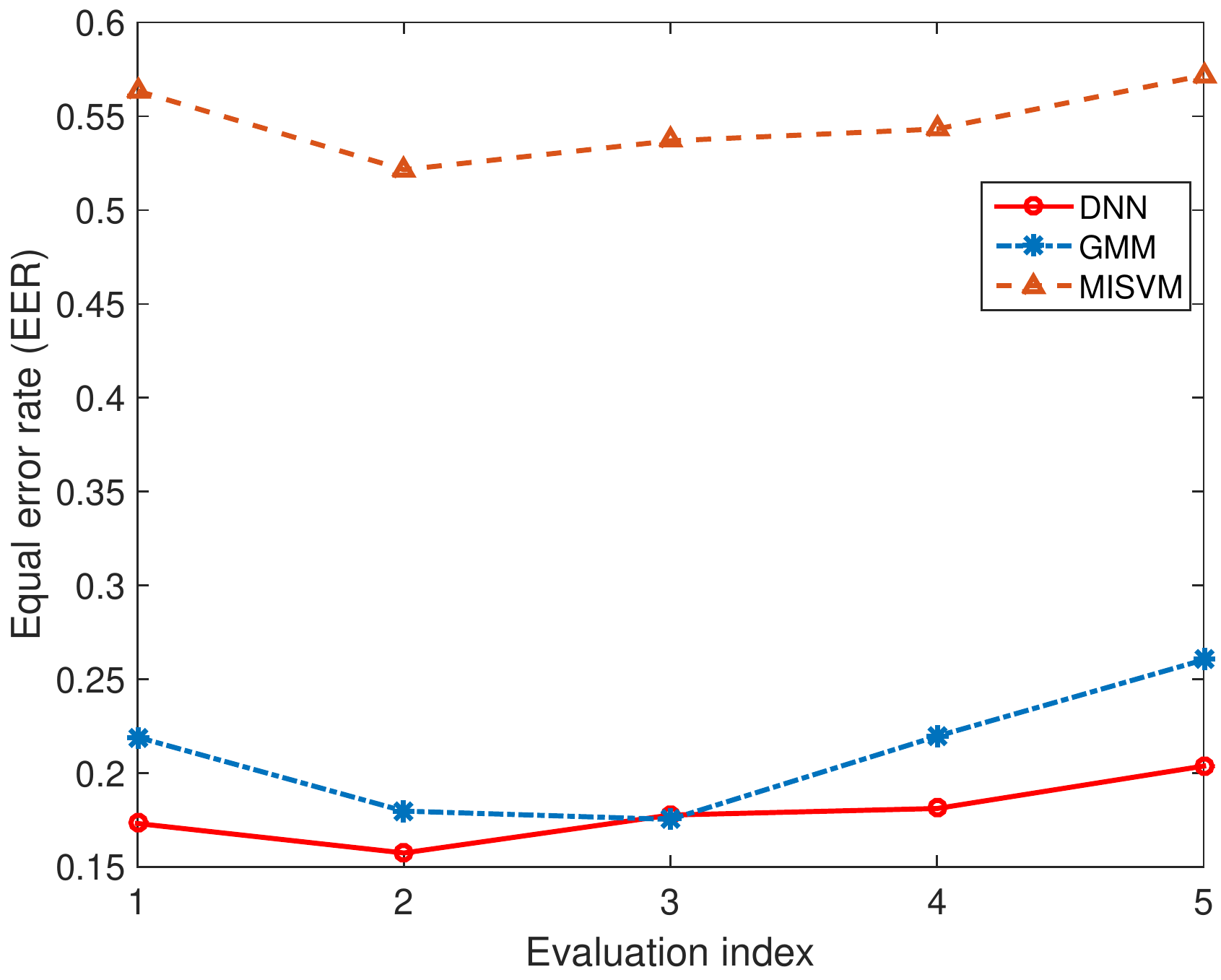} 
	\caption{Equal error rates obtained using the proposed fully DNN based approach and the two baselines, namely GMM and MI-SVM across five evaluations folds of the development set.}
	\label{fig:dnn_gmm_misvm}
\end{figure}

Figure \ref{fig:dnn_gmm_misvm} shows the results obtained using
our approach and two baselines.
Fully DNN-based approach outperforms
the two baselines across the five-fold evaluations.
This may be because of the following two main reasons:
First, our proposed approach can well utilize
the long-term temporary information instead of treating
those information independently. Second, it can
map the whole audio features sequence
into a multi-tag vector by working
as an encoding function. However, GMM and SVM based methods build the models only on single instances. The contextual information and the potential relationship among different tags were not well utilized.

The GMM based method yields a close performance to the
proposed method only in the third evaluation. 
We find that two of the audio event classes, namely
adult male's speech (label `m') and other identifiable sounds (label `o'),
are well identified in this fold evaluation.
This case is probably because the acoustic characteristics  
and their variations of the two event classes
in the evaluation data can match with the trained models.  
The use of MI-SVM does not yield competitive performances
in comparison with our proposed approach and the GMM-based baseline.
This is because MI-SVM, actually working as a
discriminative learning, is more sensitive to the quantity and quality
of the used training data. Furthermore, MI-SVM does not use the long contextual information.

\begin{table}[t]
\centering
\caption{Average EER among the proposed fully DNN method, GMM and MI-SVM methods, for each event across five-fold evaluations of the development set.}
\begin{tabular}{|c|c|c|c|}
\hline
 Various tag & Proposed DNN  & GMM & MI-SVM\\ \hline
 b & 0.0868 & \textbf{0.0755} & 0.1672 \\ 
 c & \textbf{0.1686} & 0.2107 & 0.6466\\
 f & \textbf{0.2409} & 0.3037 & 0.7626\\
 m & \textbf{0.1943} & 0.2847 & 0.7046\\
 o & \textbf{0.2867} & 0.2903 & 0.7303\\
 p & \textbf{0.2197} & 0.2613 & 0.6724\\
 v & 0.0530 & \textbf{0.0484} & 0.1481\\ \hline
 Average & \textbf{0.1785} & 0.21 &  0.5474\\
\hline
\end{tabular}
\label{tab:comparison}
\end{table}

For a further comparison, Table \ref{tab:comparison} shows the detailed 
performances obtained using our approach and the two baselines on each audio tag.
We can easily find that the use of the fully-DNN based approach
yields great improvements over the two baselines across 
all of the seven audio tags. Compared with the GMM method, the proposed fully DNN method could get similar performance on tag `b' and `v', but it can significantly outperform the competing counterparts on some difficult tags. On average, the proposed DNN method could get a relative 15\% improvement by contrasting with the GMM baseline.

\begin{table}[h]
	\begin{center}
		\begin{tabular}{|c|c|c|}
			\hline
			System & Proposed method & DCASE2016 Baseline \\
			\hline
			EER & 19.0\% & 20.9\% \\
			\hline
		\end{tabular}
	\end{center}
	\caption{EER (\%) for the final evaluation set.}
	\label{tab:eer_evaluation}
\end{table}
Table \ref{tab:eer_evaluation} presents the final EER for the evaluation set. The final DNN model was trained with the whole segments of the development set. Note that the proposed method achieve only 19.5\% EER if the DNN was trained on Fold1 only (as on the DCASE2016 Task4 website).

\section{Conclusions}\label{sec:conclusions}
 
In this paper we have presented to use a fully-DNN based
approach to handle audio tagging with weak labels, in the sense that only the chunk-level instead of the frame-level labels are available. This fully DNN is regarded as an encoding function to map the audio features sequence to a multi-tag vector in a regression way. To extract robust high-level features, a deep pyramid structure was designed to reduce most of the non-correlated interfering features while keeping the highly related features. The dropout and background noise aware training methods were adopted to further improve its generalization capacity for new recordings in unseen environments.
We tested our approach on the dataset of the Task4 of the DCASE 2016 challenge,
and obtained significant improvements over two baselines, namely GMM and MI-SVM. Compared with the official GMM-based baseline system given in the DCASE 2016 challenge, the proposed DNN system could reduce the EER from 0.21 to 0.1785 on average. For the future work, we will use fully convolutional neural network (CNN) to extract more robust high-level features for the audio tagging task.

\bibliographystyle{IEEEtran}
\bibliography{refs}
%
%
%
%
%
%
%
%
%

\end{sloppy}
\end{document}